\documentclass{article}

\usepackage[preprint]{neurips_2024}

\usepackage[utf8]{inputenc} 
\usepackage[T1]{fontenc}    
\usepackage{hyperref}       
\usepackage{url}            
\usepackage{booktabs}       
\usepackage{amsfonts}       
\usepackage{nicefrac}       
\usepackage{microtype}      
\usepackage{xcolor}         
\usepackage{url}
\usepackage{graphicx}
\usepackage{array}
\usepackage{colortbl}
\usepackage{amssymb}        
\usepackage{pifont}         
\usepackage{geometry}
\usepackage{amsmath}
\usepackage{subcaption}    
\usepackage{caption}
\usepackage{tikz}

\usepackage{float}
\usepackage{etoolbox}
\usepackage{lscape}
\usepackage{rotating}
\usepackage{diagbox}
\usepackage{xcolor,colortbl}
\usepackage{color}

\usepackage{soul}
\usepackage{hhline} 
\usepackage{multirow}
\usepackage{mwe}    
\usepackage{algorithm}
\usepackage{algpseudocode}
\usepackage{placeins}
\usepackage{tabularx}       
\usepackage{tabu}       
\usepackage{tabularray}
\usepackage{wasysym}
\usepackage{comment}

\title{ A Contextualized BERT model for Knowledge Graph Completion}

\author{%
  Haji Gul \\
  School of Digital Science, \\
  Universiti Brunei Darussalam \\
  \texttt{23h1710@ubd.edu.bn} \\
  \And
  Abul Ghani Haji Naim \\
  School of Digital Science, \\
  Universiti Brunei Darussalam \\
  \texttt{ghani.naim@ubd.edu.bn} 
  \AND
  Ajaz Ahmad Bhat \\
  School of Digital Science, \\
  Universiti Brunei Darussalam \\
  \texttt{ajaz.bhat@ubd.edu.bn} \\
}

\begin{document}

\maketitle
\begin{abstract}
Knowledge graphs (KGs) are valuable for representing structured, interconnected information across domains, enabling tasks like semantic search, recommendation systems and inference. A pertinent challenge with KGs, however, is that many entities (i.e., heads, tails) or relationships are unknown. Knowledge Graph Completion (KGC) addresses this by predicting these missing nodes or links, enhancing the graph's informational depth and utility. Traditional methods like TransE and ComplEx predict tail entities but struggle with unseen entities. Textual-based models leverage additional semantics but come with high computational costs, semantic inconsistencies, and data imbalance issues. Recent LLM-based models show improvement but overlook contextual information and rely heavily on entity descriptions. In this study, we introduce a contextualized BERT model for KGC that overcomes these limitations by utilizing the contextual information from neighbouring entities and relationships to predict tail entities. Our model eliminates the need for entity descriptions and negative triplet sampling, reducing computational demands while improving performance. Our model outperforms state-of-the-art methods on standard datasets, improving Hit@1 by 5.3\% and 4.88\% on FB15k-237 and WN18RR respectively, setting a new benchmark in KGC.

\end{abstract}
\section{Introduction}
A knowledge graph (KG) is a structured representation of entities (as nodes) and relationships (as links) that supports search, recommendation and other downstream reasoning tasks. However, KGs are often incomplete, with many entities (heads/tails) or relationships missing, limiting their utility in real-world applications \cite{ding2018}. Consequently, Knowledge Graph Completion (KGC)---predicting a missing tail entity $(h, r, ?)$, head entity $(?, r, t)$, or relationship $(h, ?, t)$ in a triplet---has become a critical research objective, with numerous methodologies proposed to tackle this issue.

Embedding-based methods, for instance, learn vector embeddings for entities and relationships from training data, but these methods struggle to generalize to \textit{unseen} entities or relationships, impairing performance in tail prediction during testing \cite{xie2016}. Recently, large language model (LLM)-based approaches for KGC have shown potential in overcoming this limitation by leveraging LLMs trained on extensive datasets to capture complex semantic relationships and generalize better to unseen entities \cite{yao2019a, wei2023, zhang2020, wei2023}. Despite these strengths, LLM-based models are computationally demanding, often overlook relation context, and depend heavily on entity descriptions and negative sampling. More recent LLM prompting-based approaches encode KGs into prompts\cite{wei2023}, but injecting all relevant facts from a KG into prompts is labor-intensive, and generic LLMs often struggle with domain-specific KGs. Additionally, textual information-based methods like NN-KGC and Sim-KGC utilize neighborhood information for KGC, but they often require entity descriptions, which may not be available in many datasets, and add computational overhead \cite{wang2022, li2024}. 
To address these limitations, we propose a Context-Aware BERT for Knowledge Graph Completion (CAB-KGC) that extracts contextual information associated with the operational relationship, its neighboring entities, and relationships associated with the head entity. This context is then integrated with the BERT model to enhance the prediction of tail entities. To summarise, this study makes the following contributions to the KG domain:
\setlength{\itemsep}{0pt}
\setlength{\parskip}{0pt}
\begin{itemize}
    \item We introduce the CAB-KGC approach to address the KGC problem, leveraging graph features of head entity context and relationship context and the BERT model. The CAB-KGC approach outperforms SOTA KGC methods.
    \item CAB-KGC eliminates reliance on entity descriptions, focusing solely on head and relationship contexts for improved predictions available in all KGs.
    \item CAB-KGC does not require negative sample training, enhancing training speed and resilience against negative sample selection.
    \item Extensive experiments on various benchmark datasets demonstrate that CAB-KGC reliably excels in tail entity prediction.
\end{itemize}
\section{Methodology}
Problem Formulation (see Table \ref{tb:mn_sm} for notations): Consider a knowledge graph $G(E, R)$ as a collection of triplets ${(h, r, t)}$, where $ h ~\epsilon ~ E ~ $ is the head entity, $ t ~ \epsilon ~ E$ is the tail entity, and $r ~ \epsilon ~ R$ represents the relationship between them, our  CAB-KGC model predicts a missing tail $t$ (represented by ?) given an incomplete triple $(h ~ , ~ r ~, ~ ?)$. 

\begin{table}[h!]
\centering
\caption{Mathematical Notations and Symbols}
\label{tb:mn_sm}
\small
\resizebox{\textwidth}{!}{  
\begin{tabular}{c l c l}
\hline
\textbf{Notation} & \textbf{Description} & \textbf{Notation} & \textbf{Description} \\
\hline
$e$ & entity or node & $r$ & relationship \\

$h$ & head entity node & $t$ & tail entity node \\

${E}$ & Entities Set & ${R}$ & Relationships Set \\

${H}_c$ & Head ($h$) or Entity context & ${R}_c$ & Relationship context\\

${R} (h), {E} (h)$ & relation and entities associated to head $h$ & $N_{T}$ & Total number of triplets\\

$p_{\theta}(t_i \mid h_i, r_i)$ & Tail $t_i$ probability given head $h_i$ and relationship $r_i$  & $\text{rank}_i$ & Rank of the true tail entity $t_i$ in the prediction\\

$\downarrow$ & Results: Lower is better & $\uparrow$ & Results: Higher is better \\
\hline
\end{tabular}
}
\end{table}

\vspace{-0.5em}
\begin{figure}[h!]
 \centering
  \includegraphics[width= 0.8\linewidth]{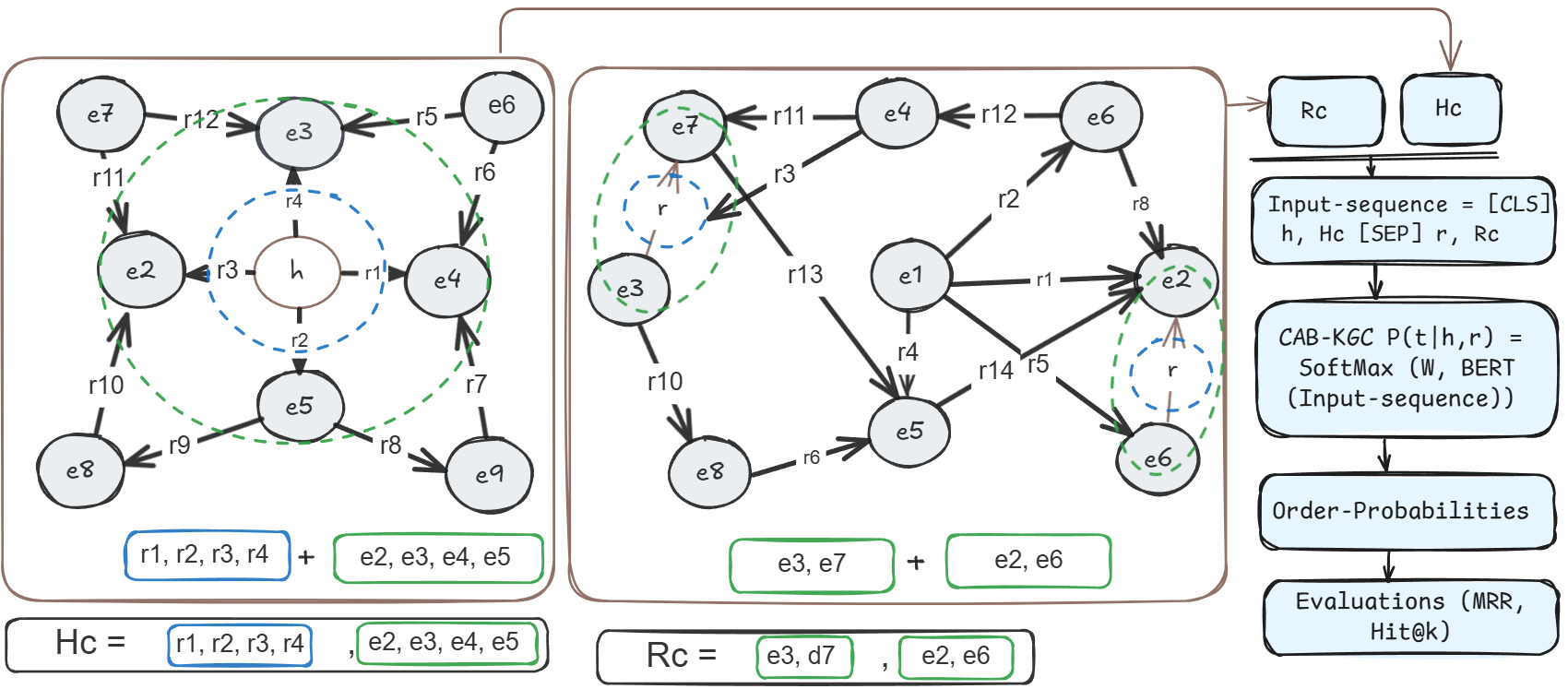} 
  \caption {A concise view of the CAB-KGC Method. Box on the left shows head context $H_c$ calculation; the middle one shows relationship context $R_c$ calculation.  $H_c$ and $R_c$ are then fed into the model pipeline shown on the right side. }
  \label{f:kgc-ca-quick}
\end{figure}
\vspace{-0.5cm}
\par
\noindent

 Figure \ref{f:kgc-ca-quick} provides an overview of the CAB-KGC model. It predicts the tail entity $t$ given a head  $h$ and a relationship $r$, in the following steps: 
\begin{enumerate}

\item  \textbf{Extract Head Context} ${H}_c$:
To extract the contextual information for the head i.e. ${H}_c$, we first identify the relationships $r$ that are associated with the head entity $h$, i.e., ${R}(h)$.  If  $k$ relationships are associated with the head $h$ from  the set $R$ of all relationships $r_i$ in the graph $G$, then:
\vspace{-5pt}
\begin{equation}
 \mathcal{R}(h) = A_{i=1}^{k}\left(\{r_i \mid (h, r_i, e_j) \in T, \, e_j \in {E}\}\right)
\end{equation}
Next, we find the entities $e$ that are neighbours (have a direct connection) with the head entity $h$, i.e., ${E}(h)$ using the identified relationships ${R}(h)$.These neighbour entities can be mathematically expressed as:

\begin{equation}
     \mathcal{E}(h) = A_{i=1}^{m}\left(\{e_i \mid (h, r_j, e_i) \in T, \, r_j \in {R}\}\right)
\end{equation}

The head context $H_c$ is then calculated as the union of the connected relationships ${R}(h)$ and the neighbour entities ${E}(h)$, as shown below in Equation \ref{eq:hc}.

\begin{equation}
{H}_c = \left( \mathcal{R} (h) \right) \cup \left( \mathcal{E} (h) \right)
\label{eq:hc}
\end{equation}

\item \textbf{Extract Relationship Context} ${R}_c$: To acquire the relationship context ${R}_c$, we identify all the entities associated with the operational relationship $r$ in the knowledge Graph $G$. $R_c$ is given as:
\setlength{\belowdisplayskip}{0pt}
\begin{equation}
    R_c = A_{i,j=1}^{l}\left(\{e_i, e_j \mid (e_i, r, e_j) \in T\}\right)
    \label{eq:rc}
\end{equation}
\par
\noindent
\item
\textbf{Prepare Input Sequence for BERT Classifier}: The contextual information extracted in the above steps forms the input to BERT. Specifically, the input sequence contains h, ${H}_c$ from  Equation \ref{eq:hc}, $r$, and ${R}_c$ from Equation \ref{eq:rc}, as shown below:
\begin{equation}
    \text{Input Sequence}   =[CLS] ~ h, ~ {H}_c ~ [SEP] ~ r, ~{R}_c
\end{equation}

where [CLS] is BERT's classifier token and [SEP] is the separator token.

\item 
\textbf{Predict and train with BERT Classifier}:
 A classification layer is added on top of the BERT model, which  aims to classify the tail entity $(h ~ , ~ r ~, ~ ?)$. Once the BERT classifier receives the input, it processes it through various transformer layers, provides a contextualized representation of each token and uses that to classify the input.  The classifier model predicts the tail entity by employing a softmax function over the output embedding to calculate the probability for all the available tail entities. The input-output description of the model is given as:
\begin{equation}
    P(t \mid h, r) = \text{softmax}(W \cdot \text{\text{ BERT(Input Sequence)}})
\end{equation}

Where \( W \) is a learned weight matrix. Putting the above equations together, the CAB-KGC model can be expressed as:
\vspace{-0.5cm}
\begin{equation}
    \text{CAB-KGC}(t \mid h, r) = \text{softmax}(W \cdot \text{ BERT}(h,{H}_c, r,{R}_c))
\end{equation}
The CAB-KGC model is trained using cross-entropy loss, which compares the probability distribution of the predicted label with the true label for the tail entity. The cross-entropy loss is given by:
\vspace{-0.5cm}
\setlength{\belowdisplayskip}{0pt}
\begin{equation}
    {L} = - \sum_{i=1}^{N} y_i \log P(t_i \mid h, r)g
\end{equation}
\par
\noindent
In this equation the one-hot encoded true label for the tail object \( t_i \) is indicated as \( y_i \). The predicted probability for the true tail entity could be denoted as \( P(t_i \mid h, r) \), where $h$ is the head and $r$ is the relation.
\end{enumerate}

\subsection{Experiments Setup}
\label{ex:det}
\textbf{Datasets:} We assessed the proposed CAB-KGC model on various commonly used KG datasets. 
These datasets are briefly explained here:
\begin{itemize}

\item FB15k-237 \cite{bollacker2008} is an updated version subset of the FB15k dataset, where the inverse triplets have been removed to increase the difficulty of the KGC. It has 14541 unique entities and 237 relationships.

\item WN18RR \cite{miller1995} is the subset of WN18, where the reverse triplets are removed, making it more complex for the models to incorporate the problem of KGC.
\end{itemize}

\textbf{Hyperparameters:} The experiments used a batch size of 16 and a learning rate of 5e-5, Adam as the optimizer and cross-entropy as the loss function. The experiments were accomplished on an NVIDIA GeForce RTX 3090 GPU with 24 GB of memory. Training for the CAB-KGC model was halted once evaluation metrics stabilized to the third decimal place.

\textbf{Evaluation:} Various standard evaluation metrics in KGC,  as given in Equation \ref{eq:eva}, such as MRR, and Hit@k, are utilized to assess the performance of the proposed method and other state-of-the-art approaches. 
\vspace{-0.5cm}
\begin{equation}
\label{eq:eva}
\text{MRR} = \frac{1}{N} \sum_{i=1}^{N} \frac{1}{\text{rank}_i} \quad ; \quad  
\text{Hits@k} = \frac{1}{N} \sum_{i=1}^{N} \mathbf{1}(\text{rank}_i \leq k)
\end{equation}
where \(\text{rank}_i\) is the correct entity rank position in the descending order sorted list of predicted scores for the \(i\)-th triplet. The function \(\mathbf{1}(\text{rank}_i \leq k)\) is a ranking function that outputs one if the true entity is ranked within the top \(k\) predictions and 0 otherwise.
\subsection{Results}
Our CAB-KGC approach shows superior results on the FB15k-237 dataset. CAB-KGC's significant performance is its Hits@1 score of 0.322, which improves SOTA by almost 5.3\%, showing a superior ability to rank accurate entities in the first place. It obtains a Hits@3 score of 0.399  and improves by 0.5\%, notably above other models, indicating that CAB-KGC reliably predicts relevant entities within the top 3 ranks. The CAB-KGC method performed well on the WN18RR dataset, getting an MRR of 0.685, which is an improvement of 1.2\%  over SOTA models and a Hits@1 of 0.637, an improvement of 4.88\%, outperforming state-of-the-art methods. All the results are reported in Table \ref{tab:results}. Note that we have excluded results from KICGPT\cite{wei2023} during comparison for reasons: (a) the model is prompt-based and not trainable, (b) its performance is highly dependent on the underlying LLM's knowledge base and (c) large KGs cannot be injected as prompts to this model. Furthermore, injecting all relevant facts from different KGs into prompts is labor-intensive, and underlying LLM will oftern struggle with domain-specific KGs when it does not contain enough relevant knowledge.
\setlength{\belowdisplayskip}{0pt}
\begin{table}[h!]
\centering
\caption{\footnotesize Comparison of the proposed and baseline methods on the datasets FB15k-237 and WN18RR. The optimal outcome for each metric is highlighted in bold, while the second-best result is underlined. The circle symbol $\Circle$ denotes that the results have been extracted from the study conducted by Wei et al. \cite{wei2024}, while the symbol $\Box$ indicates that the results have been extracted from the study conducted by Yao et al. in \cite{yao2019a}.\\ } 
\label{tab:results}
\footnotesize 
\renewcommand{\arraystretch}{1.1} 
\resizebox{\textwidth}{!}{%
\begin{tabular}{lp{1.5cm}p{1.5cm}p{1.5cm}p{1.5cm}p{1.5cm}p{1.5cm}}
\toprule
\textbf{Dataset} & \multicolumn{3}{c}{\textbf{FB15k-237}} & \multicolumn{3}{c}{\textbf{WN18RR}} \\ 
\cmidrule(lr){2-4} \cmidrule(lr){5-7}
\textbf{Methods} & \textbf{MRR $\uparrow$} & \textbf{Hits@1 $\uparrow$} & \textbf{Hits@3 $\uparrow$} & \textbf{MRR $\uparrow$} & \textbf{Hits@1 $\uparrow$} & \textbf{Hits@3 $\uparrow$} \\
\midrule
\textbf{\textit{Embedding-Based Methods}} & & & & & & \\
RESCAL \cite{nickel2011} $\Circle$      & \underline{0.356} & 0.266 & 0.390 & 0.467 & 0.439 & 0.478 \\
TransE \cite{bordes2013}   $\Circle$     & 0.279 & 0.198 & 0.376 & 0.243 & 0.043 & 0.441 \\
DistMult \cite{yang2014}  $\Circle$   & 0.241 & 0.155 & 0.263 & 0.430 & 0.390 & 0.440 \\
ComplEx \cite{trouillon2016}  $\Circle$    & 0.247 & 0.158 & 0.275 & 0.440 & 0.410 & 0.460 \\
RotatE \cite{sun2019}  $\Circle$     & 0.338 & 0.241 & 0.375 & 0.476 & 0.428 & 0.492 \\
TuckER \cite{wang2019}  $\Circle$     & \textbf{0.358}& \underline{0.266} & 0.394 & 0.470 & 0.443  & 0.482 \\
CompGCN \cite{vashishth2020} $\Circle$     & 0.355 & 0.264 & \underline{0.390} & 0.479 & 0.443 & 0.494 \\
HittER \cite{chen2020}  $\Circle$     & 0.344 & 0.246 & 0.380 & 0.496 & 0.449 & 0.514 \\
HAKE \cite{zhang2022}  $\Circle$       & 0.346 & 0.250 & 0.381 & 0.497 & 0.452 & 0.516 \\
\midrule
\textbf{\textit{Text-and Description-Based Methods}}          & & & & & & \\
Pretrain-KGE \cite{zhang2020} $\Circle$    & 0.332 &   -   &  -    & 0.235 & - & - \\
StAR \cite{wang2021}  $\Circle$           & 0.263 & 0.171 & 0.287 & 0.364 & 0.222 & 0.436 \\
MEM-KGC (w/o EP) \cite{choi2021} $\Circle$& 0.339 & 0.249 & 0.372 & 0.533 & 0.473 & 0.570 \\
MEM-KGC (w/ EP) \cite{choi2021} $\Circle$ & 0.346 & 0.253 & 0.381 & 0.557 & 0.475 & 0.604 \\
SimKGC \cite{wang2022} $\Circle$          & 0.333  & 0.246 & 0.363 & 0.671 & 0.585 & \textbf{0.731} \\
NNKGC \cite{liyang2023} $\Circle$  & 0.338  & 0.252  & 0.365 & \underline{0.674} & \underline{0.596} & \underline{0.722} \\
\midrule
\textbf{\textit{LLM-Based Methods}}         & & & & & & \\
ChatGPTzero-shot \cite{zhang2020}  $\Box$      & - & 0.237  & - & - & 0.190 & - \\
ChatGPTone-shot \cite{zhang2020} $\Box$        & - & 0.267  & - & - & 0.212 & - \\
KICGPT \cite{zhang2020}   $\Box$          & 0.410 & 0.321 & 0.430 & 0.564 & 0.478 & 0.612 \\
\midrule
\textbf{\textit{Proposed}}                   & & & & & & \\
Proposed CAB-KGC & 0.350 & \textbf{0.322} & 0.399 & \textbf{0.685} & \textbf{0.637} & 0.687 \\
\bottomrule
\end{tabular}%
}
\end{table}



\section{Conclusion}
 The proposed CAB-KGC approach exploits the contexual information to outperform existing methods in MRR and Hit@k measures, with improvements of 5.3\% and 4.88\% over FB15k-237 and WN18RR, respectively.
\medskip

\bibliographystyle{plainnat}   
\bibliography{Main_neurips_2024}           

\end{document}